\documentclass[preprint,12pt]{elsarticle}

\usepackage{amssymb}
\usepackage{amsmath}
\usepackage{float}
\usepackage{graphicx}%
\usepackage{multirow}%
\usepackage{amsmath,amssymb,amsfonts}%
\usepackage{amsthm}%
\usepackage{mathrsfs}%
\usepackage[title]{appendix}%
\usepackage{xcolor}%
\usepackage{textcomp}%
\usepackage{manyfoot}%
\usepackage{booktabs}%
\usepackage{colortbl}
\usepackage{algorithm}%
\usepackage{algorithmicx}%
\usepackage{algpseudocode}%
\usepackage{listings}%
\usepackage{multirow}
\usepackage{tabulary}
\usepackage{changepage} 
\usepackage{soul}
\setulcolor{blue}
\usepackage{makecell} 
\usepackage{url}
\usepackage{hyperref}
\usepackage{verbatim}
\usepackage{placeins}
\usepackage{todonotes}
\usepackage{pdflscape}
\usepackage{adjustbox}
\usepackage{subcaption} 

\usepackage[normalem]{ulem}

\newcommand{\ot}[1]{\textcolor{orange}{#1}}
\newcommand{\ja}[1]{\textcolor{purple}{#1}}

\journal{Information Fusion}

\begin{document}

\begin{frontmatter}



\title{IPatch: A Multi-Resolution Transformer Architecture for Robust Time-Series Forecasting}


\author[a]{Aymane Harkati}
\affiliation[a]{organization={CCPS Laboratory, ENSAM, Hassan II University},
            city={Casablanca},
            postcode={20670}, 
            country={Morocco}}
            
\author[b]{Moncef Garouani}
\affiliation[b]{organization={IRIT, UMR 5505 CNRS, Université Toulouse Capitole},
            city={Toulouse},
            postcode={31000}, 
            country={France}}

\author[c]{Olivier Teste}
\affiliation[c]{organization={IRIT, UMR 5505 CNRS, Université Toulouse, UT2J},
            city={Toulouse},
            postcode={31000}, 
            country={France}}

\author[b]{Julien Aligon}

\author[a]{Mohamed Hamlich}

\begin{abstract}

Accurate forecasting of multivariate time series remains challenging due to the need to capture both short-term fluctuations and long-range temporal dependencies. Transformer-based models have emerged as a powerful approach, but their performance depends critically on the representation of temporal data. Traditional \textit{point-wise} representations preserve individual time-step information, enabling fine-grained modeling, yet they tend to be computationally expensive and less effective at modeling broader contextual dependencies, limiting their scalability to long sequences. \textit{Patch-wise} representations aggregate consecutive steps into compact tokens to improve efficiency and model local temporal dynamics, but they often discard fine-grained temporal details that are critical for accurate predictions in volatile or complex time series. We propose IPatch, a multi-resolution Transformer architecture that integrates both point-wise and patch-wise tokens, modeling temporal information at multiple resolutions. Experiments on 7 benchmark datasets demonstrate that IPatch consistently improves forecasting accuracy, robustness to noise, and generalization across various prediction horizons compared to single-representation baselines.

\end{abstract}



\end{frontmatter}

\section{Introduction}
\label{sec1}


Time-series forecasting has increasingly adopted Transformer architectures, whose self-attention mechanism enables the modeling of long-range and heterogeneous temporal dependencies \cite{su_systematic_2025}. A central design aspect in adapting these models to temporal data is the choice of input representation (point-wise\cite{haoyietal-informerEx-2023} or patch-wise\cite{Yuqietal-2023-PatchTST}). Traditional \textit{point-wise} encodings, in which each time step is treated as an individual token, preserve full temporal resolution and allow the model to capture fine-grained local variations \cite{25_years_of_time_series_forecasting}. This property is particularly important in settings characterized by volatility or rapid fluctuations. However, point-wise representations incur substantial computational costs for long input sequences and offer limited capacity to capture broader temporal structure in the self-attention mechanism\,\cite{DLinear_zeng}. This reduces their scalability and effectiveness for long-horizon forecasting \cite{haoyietal-informerEx-2023}.

To improve scalability and incorporate localized contextual information, recent Transformer-based architectures have adopted \textit{patch-wise} representations \cite{Yuqietal-2023-PatchTST}. In this approach, consecutive observations are aggregated into compact tokens, effectively shortening the input sequence and enabling the model to capture short-range temporal patterns more efficiently. Patch-based encodings have demonstrated strong empirical performance and have been widely adopted, including in large-scale forecasting frameworks such as TimeLLM \cite{jin2024timellm}. Nevertheless, aggregating multiple time steps within each patch inevitably reduces temporal resolution, potentially discarding subtle yet informative short-term dynamics\cite{patchTCN}. This limitation becomes pronounced in irregular, noisy, or high-frequency environments where local fluctuations carry critical predictive value\cite{fredformer}.

The respective limitations of point-wise and patch-wise representations underscore their fundamentally different modeling capabilities. Point-wise encodings offer maximal temporal fidelity but scale poorly and lack the structural abstraction needed to capture broader temporal dependencies\cite{Yuqietal-2023-PatchTST}. Patch-wise encodings provide efficient and context-aware summaries but obscure fine-grained patterns\cite{patchTCN}. These characteristics indicate that the two representations capture complementary aspects of temporal dynamics and motivate the development of models capable of jointly exploiting both granular and contextual information within a unified architecture.

In this paper we introduce IPatch, a hybrid Transformer architecture that integrates point-wise and patch-wise representations within a unified framework. The design incorporates (i) a global inter-patch attention mechanism that captures coarse temporal dependencies across segments, and (ii) an intra-patch autocorrelation module that preserves fine-grained temporal structure within each patch. By jointly modeling temporal information at multiple resolutions, the architecture addresses the scaling limitations of point-wise encodings while mitigating the loss of temporal precision inherent to patch-wise aggregation. 

The remainder of this article is organized as follows. Section\,\ref{sec2} reviews related work on Transformer-based forecasting and representation strategies. Section\,\ref{sec3} describes the proposed IPatch architecture, highlighting its core components and integration strategy. Section\,\ref{experiments} reports the experimental setup and results, including ablation studies and comparisons with state-of-the-art baselines. Section\,\ref{conclusion} concludes with a summary and discussion of future research directions.

\section{Research background and state of the art}
\label{sec2}




\subsection{Time series modeling}

Time-series forecasting has evolved from classical statistical methods to deep learning architectures \cite{Lim_2021}. Traditional models, such as ARIMA \cite{din_arima_2015} and Exponential Smoothing (ETS) \cite{hyndman2002state} provide interpretable solutions for linear trends and seasonal components. While computationally efficient, they are limited in capturing nonlinear dynamics, multivariate interactions, and long-range dependencies, which restricts their applicability in complex real-world settings\,\cite{MAKRIDAKIS202054}.

Neural network-based models were introduced to overcome these limitations. Recurrent Neural Networks (RNNs) \cite{elman_finding_1990} and their gated variants LSTM and GRU \cite{yunita2025performance} improved the ability to model sequential dependencies. Hybrid architectures, such as CNN-LSTM\,\cite{CNN–LSTM}, combine convolutional layers for local feature extraction with recurrent layers for temporal modeling, reducing input complexity and enabling more effective short- to medium-term forecasting \cite{CNN-LSTM}. Despite these advances, recurrent models remain constrained by vanishing gradients, limited parallelization, and difficulty in capturing very long-horizon dependencies\,\cite{pmlr-v28-pascanu13}.

The introduction of the attention mechanism \cite{Bahdanau2014NeuralMT} marked a significant step in addressing long-range dependency challenges. By allowing the model to dynamically focus on relevant parts of the input sequence, attention mitigates the information bottleneck inherent in fixed-length context vectors of recurrent models. Transformers \cite{Attention_all_you_need}, which rely entirely on self-attention, enable parallel processing and scalable modeling of long sequences. Several Transformer-based variants have been developed for time-series forecasting, including Informer \cite{haoyietal-informerEx-2023}, Autoformer \cite{wu2021autoformer}, FEDformer \cite{zhou2022fedformer}, and DLinear \cite{DLinear_zeng}, each proposing mechanisms to address either computational complexity, periodicity modeling, or scalability. Despite these innovations, most rely on point-wise representations, encoding each timestamp as an independent token. While this preserves temporal granularity, it often fails to capture local continuity and short-term variations within the series \cite{haoyietal-informerEx-2023}.

To address the limitations of point-wise encoding, patch-based representations have been introduced. Inspired by Vision Transformers \cite{dosovitskiy2020vit}, models such as PatchTST \cite{Yuqietal-2023-PatchTST} segment the input sequence into compact patches, which are embedded and processed by attention layers. This approach reduces sequence length, improves computational efficiency, and captures local temporal dependencies. Extensions such as TimesNet \cite{wu2023timesnet} and Crossformer \cite{zhang2023crossformer} further incorporate multi-scale and cross-dimensional attention to enhance representation capacity. While patching provides scalability and structural abstraction, it inherently discards fine-grained temporal details and introduces sensitivity to patch size. Overlapping patches have been proposed to mitigate this information loss \cite{Yuqietal-2023-PatchTST}, allowing each timestamp to appear in multiple embeddings and theoretically enriching contextual representations. However, it has been demonstrated that overlapping does not necessarily lead to an improvement in performance. In some situations, it can even result in a decline in outcomes\,(see Sec.\,\ref{sec:research_problem}). 

To mitigate information loss from patching, an \textit{oversampling} strategy is proposed, where patches are extracted with overlap\,\cite{overlapping}. This design ensures that each timestamp may appear in multiple patches, theoretically enriching its contextual embedding \cite{Yuqietal-2023-PatchTST}. However, as shown in the next subsection, the benefits of overlapping are not as robust as expected and may even degrade performance in some scenarios. 

\subsection{Research problem}
\label{sec:research_problem}

\begin{sloppypar}
Patch-based representations have emerged as a popular strategy in transformers-based time series forecasting. However, despite their effectiveness at capturing global temporal dependencies, patch-based representations inherently lose fine-grained information within each segment\cite{patchTCN}. This limitation becomes particularly problematic for highly volatile or irregular time series, where subtle short-term variations are critical for accurate forecasting. A common solution is the use of \textit{overlapping} patches, inspired by n-grams in natural language processing, where each timestamp is included in multiple patches\cite{n-gram}. This redundancy is intended to enhance the model’s ability to maintain local continuity and integrate long-range temporal interactions.
\end{sloppypar}

While intuitive, overlapping patches do not consistently translate into substantial performance gains. Our empirical evaluation, detailed in Table\,\ref{tabel 1}, across 7 benchmark datasets\,\cite{haoyietal-informerEx-2023} and multiple forecasting horizons reveal that overlapping patches provide only marginal improvements. On average, the gain in Mean-Squared Error (MSE) and Mean-Absolute Error (MAE) is less than 0.01 and 0.005, respectively. In some scenarios, such as specific horizons of ILI and ETTh1 datasets, the non-overlapping performs equally well or even better. 
These results show that overlapping patches add minimal benefit, and in some scenarios, it may even blur short-term variations.

\vspace{-0.5em}

\begin{table}[h!]
\caption{Performance of PatchTST with/without overlapping patches across various datasets and horizons. Best scores are in bold blue, underlined values denote equal performance. Blue indicates gains, and red indicates losses relative to baseline.}
\label{tabel 1}%
\centering
\resizebox{0.92\textwidth}{!}{
\begin{tabular}{l|l|c|c|c|c}

\toprule
\multirow{2}{*}{Datasets}  & \multirow{2}{*}{Horizon} 
& \multicolumn{2}{c|}{PatchTST (with overlapping)} 
& \multicolumn{2}{c}{PatchTST (w/o overlapping)} \\
& & MSE & MAE & MSE & MAE \\
\midrule
\multirow{6}{*}{ILI} & 24 & \textbf{\color{blue}1.963($\blacktriangle +6.72\%$)} & \textbf{\color{blue}0.878($\blacktriangle +6.50\%$)} & 2.095 & 0.939 \\
                    & 36 &\color{red} 1.981($\blacktriangledown -1.00\%$) & \textbf{\color{blue}0.898($\blacktriangle +0.33\%$)} & \textbf{\color{blue}1.961} & 0.901 \\ 
                      & 48 & \textbf{\color{blue}2.022($\blacktriangle +0.19\%$)} & \textbf{\color{blue}0.935($\blacktriangle +1.54\%$)} & 2.026 & 0.949 \\
                      & 60 &\color{red} 1.991($\blacktriangledown -3.61\%$) & \color{red}0.924($\blacktriangledown -0.86\%$) & \textbf{\color{blue}1.919} & \textbf{\color{blue}0.916} \\
\cmidrule(lr){1-6}
\multirow{6}{*}{Electricity} & 96 & \textbf{\color{blue}0.200($\blacktriangle +2\%$)} & \textbf{\color{blue}0.288($\blacktriangle +1.73\%$)} & 0.204 & 0.293 \\
                    & 192 & \textbf{\color{blue}0.203($\blacktriangle +1.97\%$)} & \textbf{\color{blue}0.293($\blacktriangle +1.70\%$)} & 0.207 & 0.298 \\ 
                      & 336 & \textbf{\color{blue}0.194($\blacktriangle +1.03\%$)} & \textbf{\color{blue}0.284($\blacktriangle +1.40\%$)} & 0.196 & 0.288 \\
                      & 720 & \textbf{\color{blue}0.260($\blacktriangle +1.92\%$)} & \textbf{\color{blue}0.340($\blacktriangle +1.47\%$)} & 0.265 & 0.345 \\
\cmidrule(lr){1-6}
\multirow{6}{*}{Weather} & 96 & \textbf{\color{blue}0.174($\blacktriangle +1.14\%$)} & \textbf{\color{blue}0.213($\blacktriangle +1.40\%$)} & 0.176 & 0.216 \\
                    & 192 & \textbf{\color{blue}0.228($\blacktriangle +0.87\%$)} & \textbf{\color{blue}0.260($\blacktriangle +0.38\%$)} & 0.230 & 0.261 \\ 
                      & 336 & \textbf{\color{blue}0.282($\blacktriangle +0.35\%$)} & \textbf{\color{blue}0.298($\blacktriangle +0.33\%$)} & 0.283 & 0.299 \\
                      & 720 & \textbf{\color{blue}0.355($\blacktriangle +0.84\%$)} & \underline{\textbf{\color{blue}0.347}\color{blue}($\blacktriangleright +0.00\%$)} & 0.358 & \underline{\textbf{\color{blue}0.347}} \\
\cmidrule(lr){1-6}
\multirow{6}{*}{ETTh1} & 96 &\color{red} 0.377($\blacktriangledown -0.53\%$) & \underline{\textbf{\color{blue}0.397}\color{blue}($\blacktriangleright +0.00\%$)} & \textbf{\color{blue}0.375} & \underline{\color{blue}\textbf{0.397}} \\
                    & 192 & \textbf{\color{blue}0.431($\blacktriangle +1.39\%$)} & \textbf{\color{blue}0.427($\blacktriangle +0.23\%$)} & 0.437 & 0.428 \\ 
                      & 336 & \textbf{\color{blue}0.477($\blacktriangle +1.25\%$)} & \textbf{\color{blue}0.450($\blacktriangle +0.22\%$)} & 0.483 & 0.451 \\
                      & 720 & \color{red}0.484($\blacktriangledown -0.41\%$) & \color{red}0.477($\blacktriangledown -0.83\%$) & \textbf{\color{blue}0.482} & \textbf{\color{blue}0.473} \\
\cmidrule(lr){1-6}
\multirow{6}{*}{ETTh2} & 96 & \textbf{\color{blue}0.296($\blacktriangle +0.76\%$)} & \textbf{\color{blue}0.343($\blacktriangle +0.58\%$)} & 0.298 & 0.345 \\
                    & 192 & \textbf{\color{blue}0.382($\blacktriangle +0.52\%$)} & \textbf{\color{blue}0.395($\blacktriangle +0.50\%$)} & 0.384 & 0.397 \\ 
                      & 336 & \textbf{\color{blue}0.425($\blacktriangle +0.47\%$)} & \textbf{\color{blue}0.434($\blacktriangle +0.23\%$)} & 0.427 & 0.435 \\
                      & 720 & \textbf{\color{blue}0.432($\blacktriangle +0.23\%$)} & \underline{\textbf{\color{blue}0.448}\color{blue}($\blacktriangleright +0.00\%$)} & 0.433 & \underline{\textbf{\color{blue}0.448}} \\
\cmidrule(lr){1-6}
\multirow{6}{*}{ETTm1} & 96 &\color{red} 0.332($\blacktriangledown -0.60\%$) & \color{red}0.367($\blacktriangledown -0.54\%$) & \textbf{\color{blue}0.330} & \textbf{\color{blue}0.365} \\
                    & 192 & \textbf{\color{blue}0.367($\blacktriangle +1.63\%$)} & \textbf{\color{blue}0.386($\blacktriangle +0.25\%$)} & 0.373 & 0.387 \\ 
                      & 336 & \textbf{\color{blue}0.410($\blacktriangle +0.97\%$)} & \underline{\color{blue}\textbf{0.407}($\blacktriangleright +0.00\%$)} & 0.414 & \underline{\color{blue}\textbf{0.407}} \\
                      & 720 & \textbf{\color{blue}0.459($\blacktriangle +0.21\%$)} & \color{red}0.440($\blacktriangledown -0.22\%$) & 0.460 & \textbf{\color{blue}0.439} \\
\cmidrule(lr){1-6}
\multirow{6}{*}{ETTm2} 
                & 96 & \textbf{\color{blue}0.176($\blacktriangle +1.70\%$)} & \textbf{\color{blue}0.261($\blacktriangle +1.14\%$)} & 0.179 & 0.264 \\
                    & 192 & \underline{\color{blue}\textbf{0.242}($\blacktriangleright +0.00\%$)} & \underline{\color{blue}\textbf{0.303}($\small\blacktriangleright +0.00\%$)} & \underline{\color{blue}\textbf{0.242}} & \underline{\color{blue}\textbf{0.303}} \\ 
                      & 336 & \textbf{\color{blue}0.300($\blacktriangle +0.33\%$)} & \textbf{\color{blue}0.339($\blacktriangle +0.29\%$)} & 0.301 & 0.340 \\
                      & 720 & \textbf{\color{blue}0.401($\blacktriangle +0.24\%$)} & \underline{\color{blue}\textbf{0.398}($\blacktriangleright +0.00\%$)} & 0.402 & \underline{\color{blue}\textbf{0.398}} \\
\cmidrule(lr){1-6}

\end{tabular}}
\end{table}

These findings reveal that redundancy through overlapping alone is insufficient to preserve local continuity in patch-based Transformers. To address this, it is essential to adopt an approach that explicitly integrates both global and local temporal modeling. By combining patch-level attention with point-wise intra-patch representations, a hybrid framework can maintain fine-grained temporal details while capturing broader contextual dependencies. This principle underlies our proposed architecture, which unifies multi-resolution temporal information to improve both accuracy and robustness in both short- and long-term horizon forecasting.

\section{Proposed approach}
\label{sec3}


We formalize the multivariate time series forecasting task as follows.  
Let $X = (x_{i,j})_{1 \leq i \leq L, \, 1 \leq j \leq M}$ denotes a multivariate time series with $M$ variables and a lookback window of length $L$, where each $x_{\bullet j} \in \mathbb{R}^{L} $ represents a  variable. The objective is to predict the future $H$ steps, $\hat{X}=  (\hat{x}_{i,j})_{ L+1 \leq i \leq L+H , 1 \leq j \leq M } $, based on the observed sequence $X$.  



\subsection{Architecture Overview}

\begin{figure}[!htbp]
\centering

    \includegraphics[width=\linewidth]{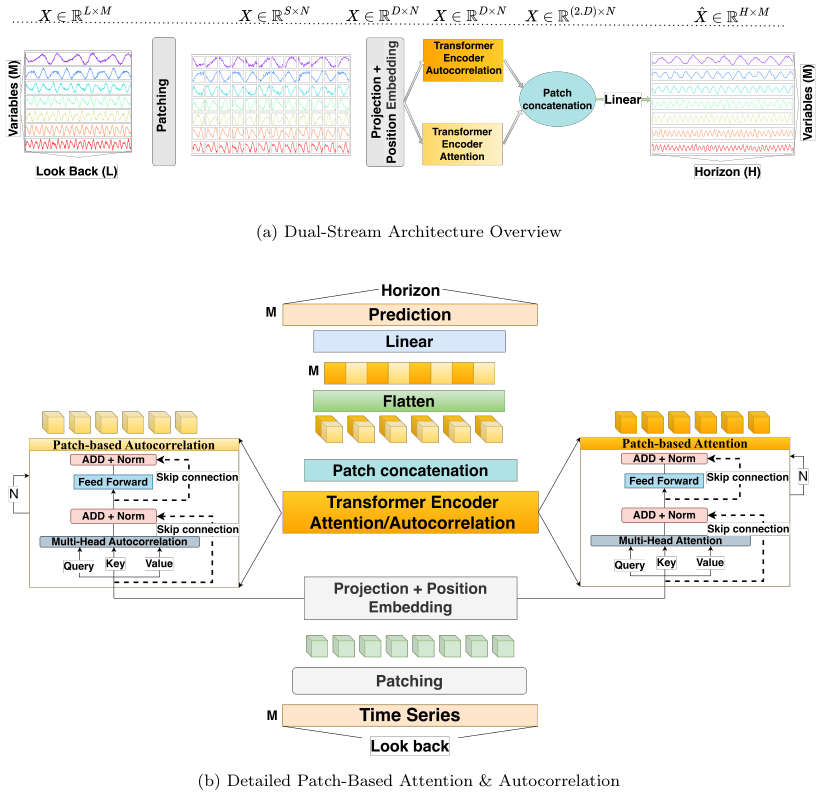}

\caption{
\textbf{(a)} The proposed dual-stream architecture. The input sequence 
$X = (x_{i,j})_{1 \leq i \leq L,\; 1 \leq j \leq M}$ is first divided into $N$ patches and combined with positional encoding. It is then passed into a \textbf{Transformer Encoder block} for patch attention and an \textbf{Autocorrelation block} for inner-patch autocorrelation. The outputs of both blocks are concatenated and flattened, then passed through a \textbf{Linear layer} to generate the final prediction 
$\hat{X} = (\hat{x}_{i,j})_{L+1 \leq i \leq L+H,\; 1 \leq j \leq M}$.
\textbf{(b)} Detailed view of the Transformer Encoder attention and the Autocorrelation block.
}
\label{IPatch_arch}

\end{figure}

Figure \ref{IPatch_arch} illustrates the overall architecture of the proposed IPatch approach. The process begins with the segmentation of the input time series into patches of uniform length, followed by the application of positional encoding to each patch to retain temporal information\,(Figure\,\ref{IPatch_arch}-a). These encoded patches are then processed through an Encoder layer, which leverages the attention mechanism to capture contextual relationships across patches, allowing each patch to integrate contextual insights from others\,(Figure\,\ref{IPatch_arch}-b, \textit{Patch-based Attention}).

To address local dependencies within individual patches, an Autocorrelation Attention block is incorporated\,(Figure\,\ref{IPatch_arch}-b, \textit{Patch-based Autocorrelation}). This block employs frequency-domain analysis to detect periodic patterns within each patch\,(point-wise), with the relatively compact patch size enhancing the efficacy of periodicity detection. The outputs from the Encoder and Autocorrelation Attention layers are subsequently concatenated, yielding enriched representations that synergistically combine inter-patch contextual information with a rolled intra-patch local details. 

    
\newpage

Formally, the input multivariate time series $X$ is segmented into $N$ patches of length $S$, with $O$ as the overlapping/non-overlapping part between patches, such that $N = \lfloor \frac{L - S}{O}\rfloor + 1$, with $1 \le O \le S < L$. The patches are defined as:  
\begin{equation}
X_{\text{patched}} = (P_{k,j})_{\, 1 \leq k \leq N , 1 \leq j \leq M}
\end{equation} 

where $P_{k,j}\in \mathbb{R}^{S\times N }$ in the $k^{th}$ patch of the variable $x_{\bullet j}$.


We use a vanilla Transformer encoder that maps the observed signals to the latent representations\,\cite{Yuqietal-2023-PatchTST}. The patches are mapped to the Transformer latent space of dimension $D$  mapping the length of patches via a linear projection, $W_{\text{proj}} \in \mathbb{R}^{D \times S}$, with the addition of a position encoding, $W_{\text{pos}} \in \mathbb{R}^{D \times N}$ to retain temporal ordering between patches,  and thus enabling the model to capture medium- and long-term trends.
Each patch is projected into a higher-dimensional embedding  using a learnable linear transformation:  
\begin{equation}
P'_{k,j} =  W_{\text{proj}}P_{k,j}  + W_{\text{pos}}, \quad  k\in [1..N],
\end{equation}  
with $P'_{k,j} \in \mathbb{R}^{D\times N }$, $W_{\text{proj}} \in \mathbb{R}^{D\times S}$,  and $W_{\text{pos}} \in \mathbb{R}^{D\times N}$.

These embeddings are fed into a Transformer Encoder, where self-attention is calculated between these patches, allowing each patch to attend to other patches, capturing global inter-patch dependencies:

\begin{equation}
Z_1(P'_{k,j}) = \text{Encoder}_{attention}(P'_{k,j}) 
\end{equation}

\begin{equation}
Z_2(P'_{k,j}) = \text{Encoder}_{autocorrelation}(P'_{k,j})
\end{equation}

with $Z_1(P'_{k,j}) \in \mathbb{R}^{D \times N}$ and $Z_2(P'_{k,j}) \in \mathbb{R}^{D \times N}$.

\subsubsection{Patch-based Attention} 

To achieve global patch-wise dependencies, we use the encoder attention block from a vanilla Transformer encoder\,\cite{Attention_all_you_need}. The set of patches $P'_{k,j}$ is represented by query $Q$, key $K$, and value $V$ matrices, in the multi-head attention, with $H_n$ as number of heads and $d_h = D /H_n$ as the dimension of each head. 
\begin{equation}
    Q_{h} =  W^{Q}_{h}P'_{k,j}, \quad K_{h} = W^{K}_{h}P'_{k,j}, \quad V_{h} = W^{V}_{h}P'_{k,j}, \quad h = 1, \dots, H_n
\end{equation}
where \( W^{Q}_{h}, W^{K}_{h}, W^{V}_{h} \in \mathbb{R}^{d_h\times D } \), are respectively the weight matrices of  \\$ Q_{h}, K_{h}, V_{h} \in \mathbb{R}^{d_h\times N} $.

The attention weights are derived by computing the scaled dot-product between the Query matrix $Q_{h}$ and the transposed Key matrix $K_{h}^\top$, followed by normalization using the softmax function across the key dimension. Formally, the attention matrix $A_{h}$ is given by : 
\begin{equation}
    A_{h} = \operatorname{softmax}\left( \frac{Q_{h} K_{h}^\top}{\sqrt{d_k}}\right), \quad h= 1,\dots,H_n
\end{equation}
where $A_{h} \in \mathbb{R}^{N \times N}$. The resulting attention distribution is then applied to the Value matrix $V_{h}$ to yield the final weighted representation:
\begin{equation}
    \operatorname{Attention_{h}}(Q_{h}, K_{h}, V_{h}) = A_{h} V_{h},\quad h= 1,\dots,H_n
\end{equation}
The outputs from the $H$ attention heads are concatenated and linearly projected to form the final multi-head attention representation. The head-specific outputs are first computed and then combined for each input patch $P'_{k,j}$, according to the following formula:

\begin{equation}
\begin{split}
Z_1(P'_{k,j}) &= \text{Encoder}_{attention}(P'_{k,j}) \\
              &= \operatorname{concat}\big( \operatorname{Attention}_{h}(Q_{h}, K_{h}, V_{h}) \big), 
                 \quad h = 1, \dots, H_n
\end{split}
\end{equation}

\subsubsection{Patch's Point-wise Autocorrelation}

To compensate the loss of fine-grained details in patching, we incorporate a \textit{point-wise Autocorrelation block}. This block processes each patch independently, capturing intra-patch temporal dependencies.

Formally, the set of patches $P'_{k,j}$ is transformed into query $Q$, key $K$, and value $V$ matrices, in the multi-head attention, with $H_n$ as number of heads and $d_h = D /H_n$ as the dimension of each head. 
\begin{equation}
    Q_{h,k} =  W^{Q}_{h}P'_{k,j}, \quad K_{h,k} = W^{K}_{h}P'_{k,j}, \quad V_{h,k} = W^{V}_{h}P'_{k,j}, \quad h = 1, \dots, H_n
\end{equation}
where \( W^{Q}_{h}, W^{K}_{h}, W^{V}_{h} \in \mathbb{R}^{d_h\times D } \), are respectively the weight matrices of  \\$ Q_{h,k}, K_{h,k}, V_{h,k} \in \mathbb{R}^{d_h\times N}$. 
Fast Fourier Transform (FFT) \cite{Nussbaumer1981} is then applied to each patch:
\begin{equation}
    \hat{Q}_{h,k} = \mathcal{F}(Q_{h,k}), \quad \hat{K}_{h,k} = \mathcal{F}(K_{h,k}), \quad h = 1, \dots, H_n
\end{equation}
where $\mathcal{F}(\bullet)$ denotes a 1-D Fast Fourier Transform applied within each patch. \\
To exploit the efficiency of frequency-domain operations, we define the transformed representations:
\begin{equation}
\hat{A}_{h,k} = \hat{Q}_{h,k} \odot \hat{K}_{h,k}^*,
\end{equation}
where $\odot$ denotes element-wise multiplication and $(\cdot)^*$ is the complex conjugate \footnote{For a complex number $z = a + bi$ (where $a, b \in \mathbb{R}$ and $i = \sqrt{-1}$), its complex conjugate is $z^* = a - bi$.}. The attention map in the spatial domain can then be recovered by the inverse Fourier transform:
\begin{equation}
A_{h,k} = \mathcal{F}^{-1}(\hat{A}_{h,k}).
\end{equation}

Applying the FFT projects the query and key representations into the frequency domain, where correlations between elements becomes more prominent\,\cite{feinashley2025spectrefftbasedefficientdropin}. Repetitive or structured patterns in the input sequence are easily captured in this domain, as frequency-domain representations simplify the temporal signal. 

A set of detected periods $\{\tau_{1,k}\dots \tau_{l,k}\}$ with their corresponding weights $\{w_{1,k}\dots w_{l,k}\}$ are used to roll the values $V$, producing temporally shifted sub-series that represent repeated patterns:  

\begin{equation}
V'_{h,k} = \text{Roll}(V_{h,k}, \tau_l),  \quad h = 1 \dots H_n
\end{equation}

The state-of-the-art softmax-based autocorrelation\,\cite{wu2021autoformer} introduces non-negative weights enabling the rejection of anti-correlated dependencies thatsuppress raw frequency-domain correlations. Instead, we employ Fourier-KAN (FKAN) layers~\cite{imran2024fourierkan}, which extend the original Kolmogorov-Arnold Network (KAN)~\cite{liu2025kan} by parameterizing the univariate functions with learnable periodic Fourier basis functions. This design yields significantly greater expressiveness than spline-based KANs, enables both positive and negative contributions through sinusoidal terms, and offers more adaptive control over periodicity. 
This approach yielded better results, see ablation study (Sec\,\ref{sec:ablation}).

\begin{equation}
\begin{split}
    \operatorname{Autocorr}_{h}(Q_{h,k}, K_{h,k}, V_{h,k}) 
= 
\sum_{\ell=1}^{J} 
\operatorname{Roll}\!\big(V_{h,k}, \tau_{\ell,k}\big) 
\odot 
\operatorname{FKAN}(w_{\ell,k})
\end{split}
\end{equation}

with $\operatorname{Autocorr}_{h,k} \in \mathbb{R}^{d_h \times N }$.

After processing each head, we obtain the rolled version of each patch. The outputs of all autocorrelation blocks are then concatenated to form the final representation.

\begin{equation}
\begin{split}
Z_2(P'_{k,j}) &= \text{Encoder}_{autocorrelation}(P'_{k,j})\\
              &= \operatorname{concat} (\operatorname{Autocorr}_{h}(Q_{h,k}, K_{h,k}, V_{h,k}) ), \quad h= 1, \dots, H_n
\end{split}
\end{equation}

\subsubsection{Patch concatenation}

Outputs from the $Z_1(P'_{k,j})$ patch-based Transformer and the $Z_2(P'_{k,j})$ Autocorrelation block are concatenated along the embedding dimension:  
\begin{equation}
C_{k,j} = \operatorname{concat}(Z_1(P'_{k,j}),Z_2(P'_{k,j}))
\end{equation}  

with $, C_{k,j} \in \mathbb{R}^{2D \times N}$.

The concatenated tensor is then flattened across patches and projected to the forecasting horizon $H$:  
\begin{equation}
\hat{X}_{L+1:L+H} = W_{\text{out}} \text{Flatten}(C_{k,j})  + b_{\text{out}}
\end{equation}

where $\hat{X}_{L+1:L+H} \in \mathbb{R}^{H\times M}$, $W_{\text{out}} \in \mathbb{R}^{H \times 2DN}$ is the weight matrix of the final linear layer that projects \( \text{Flatten}(C_{k,j}) \in \mathbb{R}^{2DN \times M} \) into the horizon $H$, and $b_{\text{out}} \in \mathbb{R}^{H \times M}$ is the bias term. 
This dual-stream design allows the model to retain global context through patch-based embeddings and local precision through point-wise Autocorrelation, effectively mitigating the shortcomings of single-representation approaches.

\section{Experimental Study}
\label{experiments}

To evaluate the effectiveness of the proposed approach, extensive experiments have been performed on widely adopted benchmark datasets. All implementations were conducted using the \texttt{TFB} package\,\cite{TFB}, which ensures reproducibility and consistency in preprocessing, environment setup, and data splits across models and baselines. For each model, dataset and forecasting horizon, we performed an independent hyperparameter optimization procedure, ensuring that every model is evaluated under its best-performing configuration. The source code of the proposed IPatch is available on \footnote{\url{https://anonymous.4open.science/r/TFB_IPatch-3132/README.md}}.

\subsection{Datasets}

We evaluate the proposed architecture on 7 widely used multivariate time-series datasets from the forecasting literature \cite{haoyietal-informerEx-2023}. These datasets span diverse real-world domains and exhibit substantial heterogeneity in their temporal structures, covering stationary dynamics, strong seasonal patterns, and long-term monotonic trends.
\begin{itemize}
    \item \textbf{Electricity Transformer Temperature} collection comprises 4 subsets \textbf{ETTh1}, \textbf{ETTh2}, \textbf{ETTm1}, and \textbf{ETTm2} capturing oil temperature and load dynamics of electricity transformers. ETTh subsets are sampled hourly, while ETTm subsets offer 15-minute resolution. Each subset contains 7 variables, where both short-term fluctuations and long-term seasonality coexist.
        
    \item \textbf{Electricity Transformer Temperature\,(ETT)} dataset consists of hourly electricity consumption records from 321 customers. It is a high-dimensional dataset (321 variables), widely used to test the scalability of forecasting models. The series exhibit strong daily and weekly seasonal patterns, as well as occasional abrupt changes caused by external events such as holidays and weather anomalies.
    \item \textbf{Weather} dataset is collected from 21 meteorological stations and includes variables such as temperature, humidity, wind speed, and solar radiation. It contains over 52,000 hourly observations. Its complex temporal dynamics arises from both short-term variability (e.g., diurnal cycles) and long-term trends (e.g., seasonal climate effects).
    \item \textbf{Influenza-Like Illness\,(ILI)} dataset records the percentage of patient visits for influenza-like symptoms across 7 regions in the United States. Unlike the other datasets, it has a weekly resolution with a relatively short history (966 timesteps). Its irregular dynamics, high noise level, and small data size make it particularly challenging, providing a strong evaluation ground for long-range generalization. 
\end{itemize}


\begin{table}[h!]
    \centering
    \resizebox{\textwidth}{!}{%
    \begin{tabular}{l|cccccccc}
    \hline
    Datasets & Weather   & Electricity & ILI & ETTh1 & ETTh2 & ETTm1 & ETTm2 \\
    \hline
    Features (M) & 21   &321  & 7&  7& 7& 7 &7\\
    Timesteps & 52696 & 26304 & 966  & 17420 & 17420 & 69680  & 69680\\
    \hline
    \end{tabular}}
    \caption{Description of the datasets used in the experiments.}
    \label{tab:placeholder}
\end{table}



\subsection{Baselines}

 To assess the efficiency of the proposed approach, it is compared against a comprehensive set of state-of-the-art baselines that span different methodological families.
 The compared models include Transformer-based architectures such as PatchTST\,\cite{Yuqietal-2023-PatchTST}, FEDformer\,\cite{zhou2022fedformer}, Autoformer\,\cite{wu2021autoformer}, Informer\,\cite{haoyietal-informerEx-2023}, and Crossformer\,\cite{zhang2023crossformer}; convolutional and spectral approaches such as TimesNet\,\cite{wu2023timesnet} and MICN\,\cite{MICN}; linear models such as DLinear\,\cite{DLinear_zeng} and TimeMixer\,\cite{wang2024timemixer}; as well as decomposition-based methods including FiLM\,\cite{FiLM} and Stationary\,\cite{liu2022nonstationary}. This collection covers the dominant paradigms in modern time series forecasting, ranging from attention-based sequence models to lightweight linear forecasters and decomposition-driven designs. Some baselines previously reported in the literature such as PAREformer\,\cite{Tong2026}, could not be included in our experimental study due to the unavailability of their source code, lack of response from the authors, or the impossibility of faithfully re-implementing their architectures given missing implementation details and hyperparameters.

All models are trained and evaluated under the same experimental configuration, ensuring a fair and consistent comparison across methods. MSE  and MAE evaluations metrics are used\,:
\begin{equation}
\begin{split}
\text{MAE} = \frac{1}{n} \sum_{i=1}^{n} |y_i - \hat{y}_i| \qquad\qquad
    \text{MSE} = \frac{1}{n} \sum_{i=1}^{n} (y_i - \hat{y}_i)^2
\end{split}
\end{equation}

\subsection{Forecasting Setup}  

For all datasets, the time series are segmented into batches of size $B$, each containing a sequence of length $T = L + H$. We use a fixed look-back window of $L = 96$ time steps across all models, resulting in input tensors of shape $(B, T, M)$, where $M$ denotes the number of variables in each dataset.
For the \textit{ETT}, \textit{Electricity}, and \textit{Weather} datasets, the forecasting horizons $H$ are set to $\{96, 192, 336, 720\}$, following established experimental protocols in prior work \cite{wu2021autoformer}. For the \textit{ILI} dataset, which is shorter and sampled at a coarser temporal resolution, we adopt horizons of $\{24, 36, 48, 60\}$, consistent with standard practice in the literature  \cite{wu2021autoformer}.
This configuration enables a systematic evaluation of forecasting performance across both short-term and long-horizon prediction scenarios.


\section{Results and discussion}

\subsection{Effectiveness of IPatch}

Table \ref{tab:comparison}  provides a detailed comparison of the proposed IPatch model against a range of state-of-the-art forecasting models across multiple datasets and varying prediction horizons. Several important observations can be drawn:

First, IPatch demonstrates a consistently robust performance across all datasets and horizons, achieving either the best results in the vast majority of cases. On short-horizon forecasting tasks, such as ILI with horizons $\{24, 36, 48, 60\}$, our approach clearly dominates: it achieves the lowest MSE and MAE in all configurations, outperforming recent strong baselines such as TimeMixer\cite{wang2024timemixer} and PatchTST\cite{Yuqietal-2023-PatchTST} by a significant margin. This highlights the ability of our hybrid patch–point design to capture both fine-grained local variations and global seasonal effects, which are especially critical in sparse and noisy health-related data.

\begin{sloppypar}
Second, in large-scale industrial benchmarks such as Electricity and Weather, IPatch is not only competitive but also consistently among the top two performers across horizons $\{96, 192, 336, 720\}$. Notably, it achieves the best overall accuracy in most short- and mid-range horizons (96, 192, 336), while maintaining competitive robustness at the longest horizon of 720 steps. This stability across horizons suggests that the model does not suffer from degradation as the forecasting window increases—a challenge commonly observed in Transformer-based baselines such as Informer or Autoformer.
\end{sloppypar}

Third, for the ETT datasets ETTh1, ETTh2, ETTm1, ETTm2, which are known for their periodic yet non-stationary behaviors, IPatch consistently outperforms or matches strong baselines like TimeMixer and PatchTST. The improvements are most pronounced at longer horizons (e.g., 336 and 720), where many models struggle to maintain accuracy. This underlines the robust long-range dependency modeling achieved by integrating patch-level representations with point-wise refinement.

 When results are aggregated across all benchmarks, IPatch achieves the best score on 26 out of 28 configurations for MAE and 21 out of 28 for MSE. This frequency highlights not only peak performance but also the robustness of our approach across diverse domains, ranging from epidemiological series (ILI) to high-dimensional industrial data (Electricity, Weather) and periodic yet non-stationary behaviors (ETT).

\begin{landscape}

\begin{table*}[htbp]
\caption{Long-term Forcasting : Unified Look-back of 96, with horizon  $\in \{96, 192, 336, 720\} $ for Electricity, Weather \& ETT, ILI horizon is set $\in  \{24, 36, 48, 60\} $ Bold results are the best outcomes, underlined are second best.}
\label{tab:comparison}
\renewcommand{\arraystretch}{1.2}

 \resizebox{1.6\textwidth}{!}{%
\begin{tabular}{l|l|c c|c c|c c|c c|c c|c c|c c|c c|c c|c c|c c|c c}
\toprule
 \multicolumn{2}{c|}{\textbf{Model}} 
& \multicolumn{2}{c|}{\textbf{IPatch (Ours)}} 
& \multicolumn{2}{c|}{\textbf{TimeMixer}} 
& \multicolumn{2}{c|}{\textbf{PatchTST}} 
& \multicolumn{2}{c|}{\textbf{TimesNet}} 
& \multicolumn{2}{c|}{\textbf{Crossformer}} 
& \multicolumn{2}{c|}{\textbf{MICN}} 
& \multicolumn{2}{c|}{\textbf{FiLM}} 
& \multicolumn{2}{c|}{\textbf{DLinear}} 
& \multicolumn{2}{c|}{\textbf{FEDformer}} 
& \multicolumn{2}{c|}{\textbf{Stationary}} 
& \multicolumn{2}{c|}{\textbf{Autoformer}} 
& \multicolumn{2}{c}{\textbf{Informer}} \\\cmidrule(lr){1-26}
 \multicolumn{2}{c|}{\textbf{Metric}}  & MSE & MAE & MSE & MAE & MSE & MAE & MSE & MAE & MSE & MAE & MSE & MAE & MSE & MAE & MSE & MAE & MSE & MAE & MSE & MAE & MSE & MAE & MSE & MAE \\
\midrule
\multirow{2}{*}{\rotatebox{90}{ILI}} 
                    & 24 & \textbf{\color{blue}1.604($ \blacktriangle +9,17\%$)} & \textbf{\color{blue}0.751($ \blacktriangle +12,38\%$)} & \underline{{1.766}} & \underline{{0.844}} & 1.963 & 0.878 & 2.616 & 1.079 & 3.197 & 1.149 & 2.341 & 1.038 & 2.223 & 0.986 & 2.203 & 1.036 & 2.565 & 1.080 & 2.301 & 1.035 & 2.651 & 1.114 & 3.179 & 1.278 \\
                    & 36 & \textbf{\color{blue}1.689($ \blacktriangle +5,50\%$)} & \textbf{\color{blue}0.799($ \blacktriangle +9,38\%$)} & \underline{{1.782}} & \underline{{0.874}} & 1.981 & 0.898 & 2.030 & 0.945 & 3.653 & 1.264 & 2.339 & 1.052 & 2.155 & 0.997 & 2.323 & 1.076 & 3.099 & 1.227 & 2.109 & 1.004 & 2.651 & 1.114 & 2.788 & 1.125 \\ 
                    & 48 & \textbf{\color{blue}1.562($ \blacktriangle +19,91\%$)} & \textbf{\color{blue}0.783($ \blacktriangle +15,58\%$)} & \underline{{1.873}} & \underline{{0.905}} & 2.022 & 0.935 & 2.032 & 0.917 & 3.903 & 1.317 & 2.437 & 1.068 & 2.144 & 1.011 & 2.436 & 1.146 & 3.183 & 1.239 & 2.477 & 1.094 & 2.846 & 1.146 & 2.942 & 1.160 \\
                    & 60 & \textbf{\color{blue}1.445($ \blacktriangle +24,84\%$)} & \textbf{\color{blue}0.783($ \blacktriangle +12,26\%$)} & \underline{{1.804}} & 0.910 & 1.991 & 0.924 & 1.836 & \underline{{0.879}} & 4.695 & 1.451 & 2.260 & 0.996 & 2.003 & 0.936 & 2.305 & 1.088 & 2.706 & 1.109 & 2.429 & 1.086 & 2.716 & 1.128 & 3.024 & 1.166 \\
\cmidrule(lr){1-26}
\multirow{2}{*}{\rotatebox{90}{Electricity}} 
                    & 96 & \textbf{\color{blue}0.153($\blacktriangleright +0 \%$)} & \textbf{\color{blue}0.239($\blacktriangle +3,34 \%$)} & \textbf{\color{blue}0.153} & \underline{{0.247}} & 0.200 & 0.288 & \underline{{0.168}} & 0.272 & 0.219 & 0.314 & 0.180 & 0.293 & 0.198 & 0.274 & 0.210 & 0.302 & 0.193 & 0.308 & 0.169 & 0.273 & 0.201 & 0.317 & 0.274 & 0.368 \\
                    & 192 & \textbf{\color{blue}0.163($\blacktriangle +1,84 \%$)} & \textbf{\color{blue}0.248($\blacktriangle +3,22 \%$)} & \underline{{0.166}} & \underline{{0.256}} & 0.203 & 0.293 & 0.184 & 0.322 & 0.231 & 0.322 & 0.189 & 0.302 & 0.198 & 0.278 & 0.210 & 0.305 & 0.201 & 0.315 & 0.182 & 0.286 & 0.222 & 0.334 & 0.296 & 0.386 \\ 
                    & 336 & \textbf{\color{blue}0.182($\blacktriangle +1,64 \%$)} & \textbf{\color{blue}0.268($\blacktriangle +3,35 \%$)} & \underline{{0.185}} & \underline{{0.277}} & 0.194 & 0.284 & 0.198 & 0.300 & 0.246 & 0.337 & 0.198 & 0.312 & 0.217 & 0.300 & 0.223 & 0.319 & 0.214 & 0.329 & 0.200 & 0.304 & 0.231 & 0.443 & 0.300 & 0.394 \\
                    & 720 & \textbf{\color{blue}0.215($\blacktriangle +4,65 \%$)} & \textbf{\color{blue}0.296($\blacktriangle +4,73 \%$)} & 0.225 & \underline{{0.310}} & 0.260 & 0.340 & 0.220 & 0.320 & 0.280 & 0.363 & \underline{{0.217}} & 0.330 & 0.278 & 0.356 & 0.258 & 0.350 & 0.246 & 0.355 & 0.222 & 0.321 & 0.254 & 0.361 & 0.373 & 0.439 \\
\cmidrule(lr){1-26}
\multirow{2}{*}{\rotatebox{90}{Weather}} 
                    & 96 & \underline{{0.164}}($\blacktriangledown -0,60 \%$) & \textbf{\color{blue}0.202($\blacktriangle +3,46 \%$)} & \textbf{\color{blue}0.163} & \underline{{0.209}} & 0.174 & 0.213 & 0.172 & 0.220 & 0.195 & 0.271 & 0.198 & 0.261 & 0.195 & 0.236 & 0.195 & 0.252 & 0.217 & 0.296 & 0.173 & 0.223 & 0.266 & 0.336 & 0.300 & 0.384 \\
                    & 192 & 0.210($\blacktriangledown -0,95 \%$) & \textbf{\color{blue}0.245($\blacktriangle +2,04 \%$)} & \textbf{\color{blue}0.208} & \underline{0.250} & 0.228 & 0.260 & 0.219 & 0.261 & \underline{{0.209}} & 0.277 & 0.239 & 0.299 & 0.239 & 0.271 & 0.237 & 0.295 & 0.276 & 0.336 & 0.245 & 0.285 & 0.307 & 0.367 & 0.598 & 0.544 \\ 
                    & 336 & 0.266($\blacktriangledown -7,51 \%$) & \textbf{\color{blue}0.286($\blacktriangle +0,34 \%$)} & \underline{{0.251}} & \underline{0.287} & 0.282 & 0.298 & \textbf{\color{blue}0.246} & 0.337 & 0.273 & 0.332 & 0.285 & 0.336 & 0.289 & 0.306 & 0.282 & 0.331 & 0.339 & 0.380 & 0.321 & 0.338 & 0.359 & 0.395 & 0.578 & 0.523 \\
                    & 720 & 0.350($\blacktriangledown -3,14 \%$) & \textbf{\color{blue}0.341($\blacktriangleright +0 \%$)} & \textbf{\color{blue}0.339} & \textbf{\color{blue}0.341} & 0.355 & \underline{0.347} & 0.365 & 0.359 & 0.379 & 0.401 & 0.351 & 0.388 & 0.361 & 0.351 & \underline{0.345} & 0.382 & 0.403 & 0.428 & 0.414 & 0.410 & 0.419 & 0.428 & 1.059 & 0.741 \\
\cmidrule(lr){1-26}

\multirow{2}{*}{\rotatebox{90}{ETTh1}} 
                    & 96 & \underline{{0.376}}($\blacktriangledown -0,26 \%$) & \textbf{\color{blue}0.391($\blacktriangle +2,30 \%$)} & \textbf{\color{blue}0.375} & \underline{{0.400}} & 0.377 & 0.397 & 0.384 & 0.402 & 0.423 & 0.448 & 0.426 & 0.446 & 0.438 & 0.433 & 0.397 & 0.412 & 0.395 & 0.424 & 0.513 & 0.491 & 0.449 & 0.459 & 0.865 & 0.713 \\
                    & 192 & \textbf{\color{blue}0.418($\blacktriangle +2,63 \%$)} & \textbf{\color{blue}0.419($\blacktriangle +0,47 \%$)} & \underline{{0.429}} & \underline{{0.421}} & 0.431 & 0.427 & 0.436 & 0.429 & 0.471 & 0.474 & 0.454 & 0.464 & 0.493 & 0.466 & 0.446 & 0.441 & 0.469 & 0.470 & 0.534 & 0.504 & 0.500 & 0.482 & 1.008 & 0.792 \\
                    & 336 & \textbf{\color{blue}0.452($\blacktriangle +7,07 \%$)} & \textbf{\color{blue}0.439($\blacktriangle +4,32 \%$)} & \underline{{0.484}} & \underline{{0.458}} & 0.477 & 0.450 & 0.638 & 0.469 & 0.570 & 0.546 & 0.493 & 0.487 & 0.547 & 0.495 & 0.489 & 0.467 & 0.530 & 0.499 & 0.588 & 0.535 & 0.521 & 0.496 & 1.107 & 0.809 \\
                    & 720 & \textbf{\color{blue}0.456($\blacktriangle +9,21 \%$)} & \textbf{\color{blue}0.466($\blacktriangle +3,43 \%$)} & \underline{{0.498}} & \underline{{0.482}} & 0.484 & 0.477 & 0.521 & 0.500 & 0.653 & 0.621 & 0.526 & 0.526 & 0.586 & 0.538 & 0.513 & 0.510 & 0.598 & 0.544 & 0.643 & 0.616 & 0.514 & 0.512 & 1.181 & 0.865 \\
\cmidrule(lr){1-26}
\multirow{2}{*}{\rotatebox{90}{ETTh2}} 
                    & 96 & \textbf{\color{blue}0.284($\blacktriangle +1,76 \%$)} & \textbf{\color{blue}0.333($\blacktriangle +2,40 \%$)} & \underline{{0.289}} & \underline{{0.341}} & 0.296 & 0.343 & 0.340 & 0.374 & 0.745 & 0.584 & 0.372 & 0.424 & 0.322 & 0.364 & 0.340 & 0.394 & 0.358 & 0.397 & 0.476 & 0.458 & 0.346 & 0.388 & 3.755 & 1.525 \\
                    & 192 & \textbf{\color{blue}0.355($\blacktriangle +4,78 \%$)} & \textbf{\color{blue}0.382($\blacktriangle +2,61 \%$)} & \underline{{0.372}} & \underline{{0.392}} & 0.382 & 0.395 & 0.402 & 0.414 & 0.877 & 0.656 & 0.492 & 0.492 & 0.404 & 0.414 & 0.482 & 0.479 & 0.429 & 0.439 & 0.512 & 0.493 & 0.456 & 0.452 & 5.602 & 1.931 \\
                    & 336 & \underline{{0.406}}($\blacktriangledown -4,92 \%$) & \underline{{0.418}}($\blacktriangledown -0,95 \%$) & \textbf{\color{blue}0.386} & \textbf{\color{blue}0.414} & 0.425 & 0.434 & 0.452 & 0.452 & 1.043 & 0.731 & 0.607 & 0.555 & 0.435 & 0.445 & 0.591 & 0.541 & 0.496 & 0.487 & 0.552 & 0.551 & 0.482 & 0.486 & 4.721 & 1.835 \\
                    & 720 & \textbf{\color{blue}0.410($\blacktriangle +0,48 \%$)} & \textbf{\color{blue}0.431($\blacktriangle +0,69 \%$)} & \underline{{0.412}} & \underline{{0.434}} & 0.432 & 0.448 & 0.462 & 0.468 & 1.104 & 0.763 & 0.824 & 0.655 & 0.447 & 0.458 & 0.839 & 0.661 & 0.463 & 0.474 & 0.562 & 0.560 & 0.515 & 0.511 & 3.647 & 1.625 \\
\cmidrule(lr){1-26}
\multirow{2}{*}{\rotatebox{90}{ETTm1}} 
                    & 96 & \textbf{\color{blue}0.304($\blacktriangle +5,26 \%$)} & \textbf{\color{blue}0.341($\blacktriangle +4,69 \%$)} & \underline{{0.320}} & \underline{{0.357}} & 0.332 & 0.367 & 0.338 & 0.375 & 0.404 & 0.426 & 0.365 & 0.387 & 0.353 & 0.370 & 0.346 & 0.374 & 0.379 & 0.419 & 0.386 & 0.398 & 0.505 & 0.475 & 0.672 & 0.571 \\
                    & 192 & \textbf{\color{blue}0.352($\blacktriangle +2,55 \%$)} & \textbf{\color{blue}0.369($\blacktriangle +3,25 \%$)} & \underline{{0.361}} & \underline{{0.381}} & 0.367 & 0.386 & 0.374 & 0.387 & 0.450 & 0.451 & 0.403 & 0.408 & 0.389 & 0.387 & 0.382 & 0.391 & 0.426 & 0.441 & 0.459 & 0.444 & 0.553 & 0.496 & 0.795 & 0.669 \\
                    & 336 & \textbf{\color{blue}0.380($\blacktriangle +2,63 \%$)} & \textbf{\color{blue}0.392($\blacktriangle +3,06 \%$)} & \underline{{0.390}} & \underline{{0.404}} & 0.410 & 0.407 & 0.410 & 0.411 & 0.532 & 0.515 & 0.436 & 0.431 & 0.421 & 0.408 & 0.415 & 0.415 & 0.445 & 0.459 & 0.495 & 0.464 & 0.621 & 0.537 & 1.212 & 0.871 \\
                    & 720 & \textbf{\color{blue}0.436($\blacktriangle +4,12 \%$)} & \textbf{\color{blue}0.424($\blacktriangle +4,00 \%$)} & \underline{{0.454}} & \underline{{0.441}} & 0.459 & 0.440 & 0.478 & 0.450 & 0.666 & 0.589 & 0.489 & 0.462 & 0.481 & \underline{{0.441}} & 0.473 & 0.451 & 0.543 & 0.490 & 0.585 & 0.516 & 0.671 & 0.561 & 1.166 & 0.823 \\
\cmidrule(lr){1-26}
\multirow{2}{*}{\rotatebox{90}{ETTm2}} 
                & 96 & \textbf{\color{blue}0.170($\blacktriangle +2,94 \%$)} & \textbf{\color{blue}0.250($\blacktriangle +3,20 \%$)} & \underline{0.175} & \underline{0.258} & 0.176 & 0.261 & 0.187 & 0.267 & 0.287 & 0.366 & 0.197 & 0.296 & 0.183 & 0.266 & 0.193 & 0.293 & 0.203 & 0.287 & 0.192 & 0.274 & 0.255 & 0.339 & 0.365 & 0.453
\\
                    & 192 & \textbf{\color{blue}0.237($\blacktriangleright +0 \%$)} & \textbf{\color{blue}0.296($\blacktriangle +1,01 \%$)} & \textbf{\color{blue}0.237} & \textbf{\color{blue}0.299} & 0.242 & 0.303 & 0.249 & 0.309 & 0.414 & 0.492 & 0.284 & 0.361 & \underline{0.248} & \underline{0.305} & 0.284 & 0.361 & 0.269 & 0.328 & 0.280 & 0.339 & 0.281 & 0.340 & 0.533 & 0.563
\\ 
                      & 336 & \textbf{\color{blue}0.297($\blacktriangle +0,33 \%$)} & \underline{0.366}($\blacktriangledown -7,10 \%$) & \underline{0.298} & \textbf{\color{blue}{0.340}} & 0.300 & 0.339 & 0.321 & 0.351 & 0.597 & 0.542 & 0.381 & 0.429 & 0.309 & 0.343 & 0.382 & 0.429 & 0.325 & 0.366 & 0.334 & 0.361 & 0.339 & 0.372 & 1.363 & 0.887 
\\
                      & 720 & \underline{0.396}($\blacktriangledown -1,26 \%$) & \textbf{\color{blue}0.392($\blacktriangle +1 ,02\%$)} &  \textbf{\color{blue}0.391} & \underline{0.396} & 0.401 & 0.398 & 0.408 & 0.403 & 1.730 & 1.042 & 0.549 & 0.522 & 0.410 & 0.400 & 0.558 & 0.525 & 0.421 & 0.415 & 0.417 & 0.413 & 0.433 & 0.432 & 3.379 & 1.338\\
\bottomrule
\end{tabular}
}

\end{table*}
\end{landscape}

\begin{figure}[h]
    \centering
        \includegraphics[width=1\textwidth]{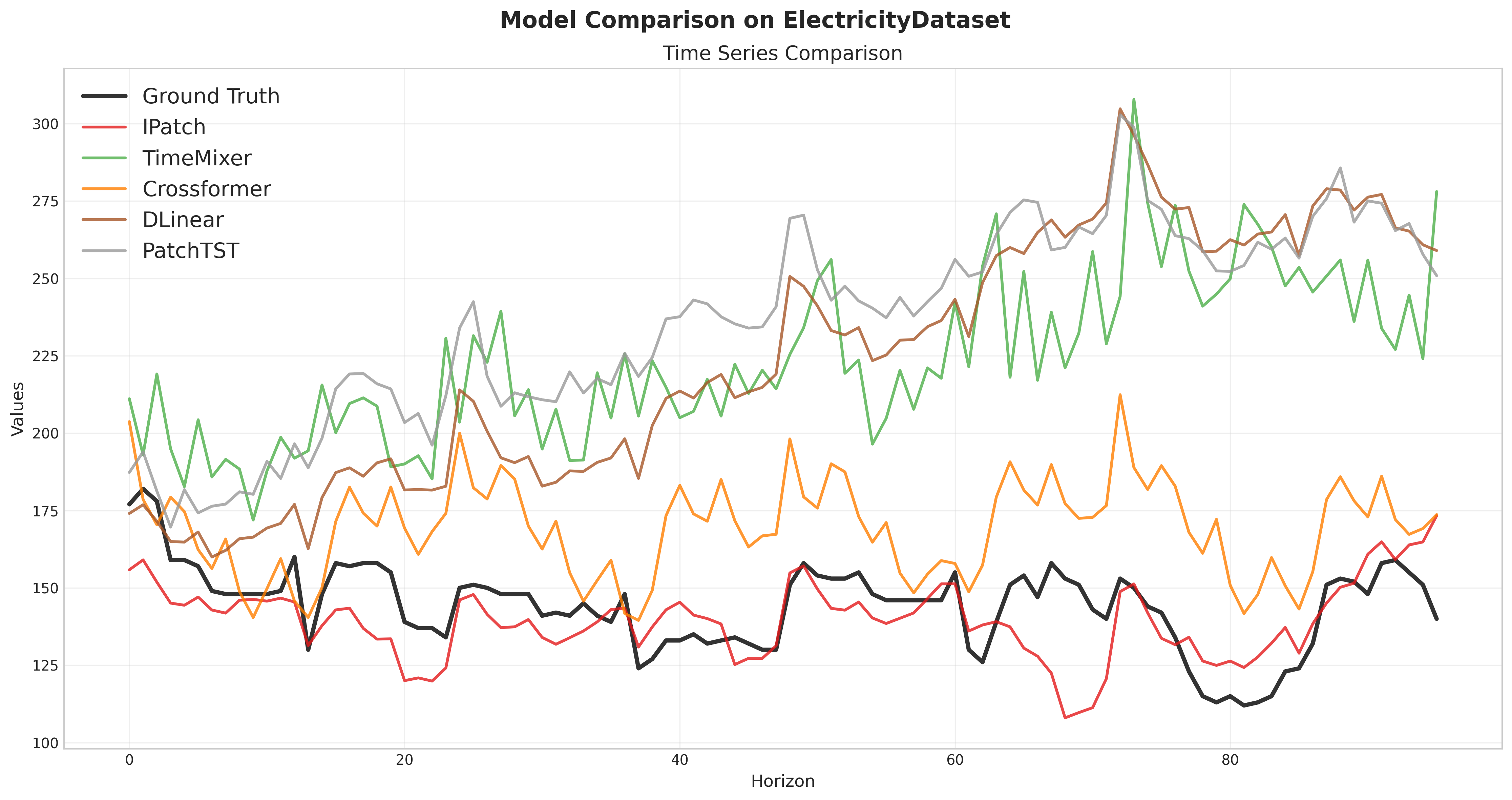} \\
    \caption{Visual comparison of 96-step forecasts on the Electricity dataset. IPatch (red) tracks the ground truth (black) with markedly higher fidelity than PatchTST, TimeMixer, Crossformer, and DLinear.}
    \label{fig:electricity_model_comparisons}
\end{figure}

Figure\ref{fig:electricity_model_comparisons} illustrates an in depth qualitative comparison on one of the representative channels of the Electricity dataset. IPatch is close to the ground-truth sequence  in the whole prediction window (96), and it is accurate in reproducing sharp oscillations, deep troughs and complex oscillations. By contrast, TimeMixer, Crossformer, DLinear, and PatchTST are systematically deviant, and they show phase shifts, damped responses, or smoothed waveforms which are unable to reflect the rich temporal dynamics. This further highlights the greater strength and faithfulness of IPatch on non-trivial time series that are greatly volatile in the real world. Additional illustrations on other datasets are provided in the appendix.

\subsection{Efficiency of IPatch}

\begin{figure}[h!]
    \centering
    \includegraphics[width=1\textwidth]{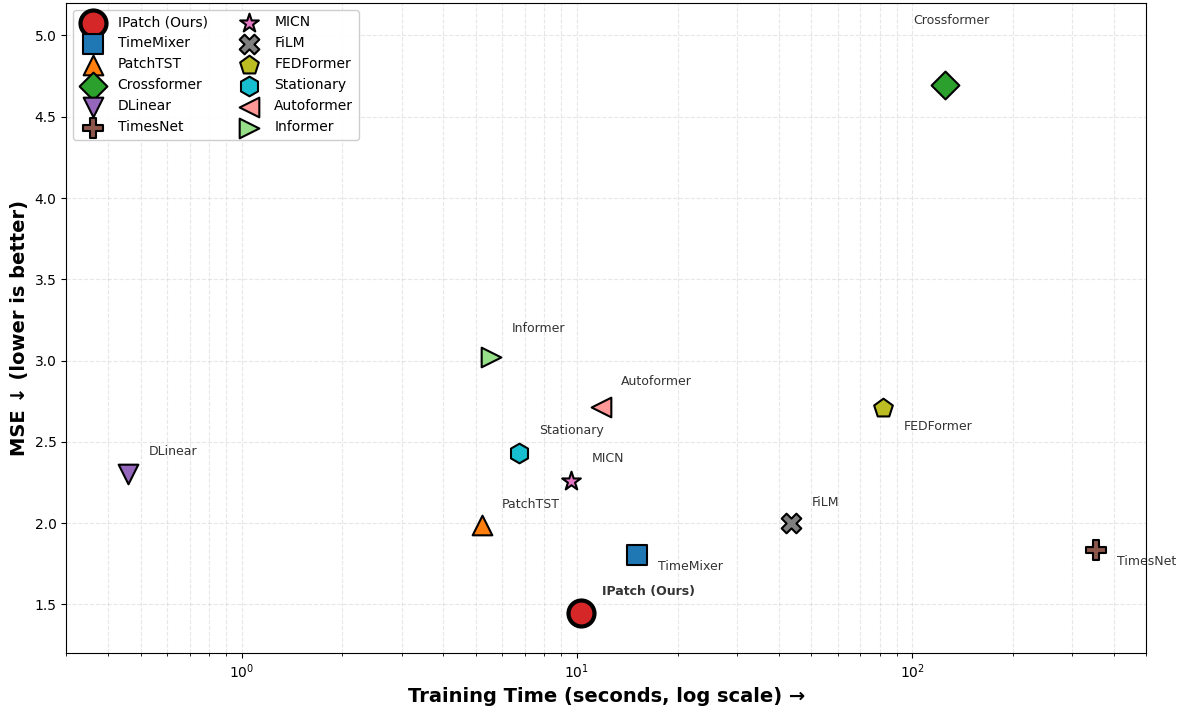}
    \caption{Predictive effectiveness vs efficiency on ILI dataset (forecast horizon 60). IPatch (red circle) offers the best performance-efficiency tradeoff among all evaluated architectures}
    \label{fig:effieciency-plot}
\end{figure}

Figure\,\ref{fig:effieciency-plot} reports the efficiency of IPatch on the ILI dataset, comparing training time (log scale) and forecasting accuracy\,(MSE) across a representative set of state-of-the-art baselines. Although lightweight linear models achieve fast training time, their speed comes at the cost of significantly degraded predictive effectiveness\,(see Table\,\ref{tab:comparison}). In particular, the DLinear baselines remain far from competitive on all the studied datasets.

IPatch, by contrast, achieves a more desirable balance: it delivers the lowest MSE among all competitors while maintaining low-latency training times. Compared with state-of-the-art Transformer-based approaches such as FEDformer, Autoformer, and Crossformer which often require one to two orders of magnitude more computation. IPatch offers a substantially more favorable effectiveness–efficiency trade-off. This advantage stems from its multi-resolution tokenization scheme, which integrates point-wise and patch-wise representations within a unified Transformer architecture. Point-wise tokens preserve the fine-grained temporal fluctuations essential for accurate short-term modeling, while patch-wise tokens encode broader temporal structures efficiently. Their joint use enables IPatch to capture both local and long-range dependencies without incurring the computational burden of full-resolution attention\,(w.r.t PatchTST).

While Figure\,\ref{fig:effieciency-plot} focuses on the ILI dataset, the same trend is consistently observed across all other benchmark datasets: IPatch sustains high predictive accuracy with significantly lower computational overhead than recent Transformer architectures, while also outperforming fast-but-weak linear baselines in predictive effectiveness.

\subsection{Ablation Study}
\label{sec:ablation}
To evaluate the contributions of the patching and point-wise autocorrelation components in the IPatch architecture, comprehensive ablation studies have been conducted, isolating the patching mechanism and autocorrelation information, as well as comparing the proposed Fourier-based activation function against the state-of-the-art softmax-based variant. These experiments provide insights into the roles of each component and the effectiveness of our design choices across the diverse datasets and prediction horizons.

\subsubsection{IPatch component complementarity}

Table \ref{tab:comparison_abelations1} compares three variants of IPatch: (1) the full IPatch model (integrating both patching and autocorrelation mechanisms), (2) Patch only\footnote{To ensure greater flexibility, and seamless integration with the overall architecture, we implemented the patch encoder from scratch.}, and (3) autocorrelation only. The results demonstrate that the full IPatch model consistently outperforms both ablated components across all datasets and horizons, as measured by Mean Squared Error and Mean Absolute Error.

On the ILI dataset, IPatch achieves the lowest errors across all horizons (e.g., MSE 1.445 and MAE 0.789 at horizon 60), significantly surpassing Patch only (MSE 1.839, MAE 0.952) and Autocorrelation only (MSE 1.711, MAE 0.870). Similar trends are observed in Weather and ETTh1, where the full model yields improvements of up to 38\% in MSE for ETTh1 at horizon 720 (0.456 vs. 0.633 (38\%) for Patch only and 0.545(19\%) for Autocorrelation only). 

The Autocorrelation only variant generally outperforms the Patch only one, particularly at longer horizons. For example, on ETTh1 at horizon 336, Autocorrelation only achieves an MSE of 0.515 compared to 0.539 for Patch only, suggesting that autocorrelation information is more effective at capturing long-term dependencies. However, both variants fall short of the full model, indicating that the patching mechanism enhances the model’s ability to process local temporal patterns, which, when combined with autocorrelation information, maximizes forecasting accuracy.

The consistent superiority of the full IPatch model underscores the complementary nature of patching and autocorrelation mechanism. For instance, on the Weather dataset at horizon 720, IPatch achieves an MSE of 0.350, compared to 0.364 for Patch only and 0.373 for autocorrelation only, highlighting how the integration of both components captures both local and global temporal dynamics effectively.

\begin{table} [h!]
\centering
\caption{Ablation study of patching and point-wise autocorrelation integration. Blue bold results are the best outcomes, underlined are second best.}
\label{tab:comparison_abelations1}
\renewcommand{\arraystretch}{1.2}

 \resizebox{0.79\textwidth}{!}{%
\begin{tabular}{l|l|c c|c c|c c}
\toprule
\multirow{2}{*}{\textbf{Datasets}} & \multirow{2}{*}{\textbf{Horizon}} 
& \multicolumn{2}{c|}{\textbf{IPatch (Ours)}} 
& \multicolumn{2}{c|}{\textbf{Patch only}} 
& \multicolumn{2}{c}{\textbf{Autocorrelation only}}  \\
& & MSE & MAE & MSE & MAE & MSE & MAE \\
\midrule
\multirow{2}{*}{ILI} 
                    & 24 &  \textbf{\color{blue}1.604} & \textbf{\color{blue}\color{blue}0.751} &  1.977 & 0.873 & \underline{1.790} & \underline{0.805}\\
                    & 36 & \textbf{\color{blue}1.689} & \textbf{\color{blue}0.799} & 1.886 & 0.882 & \underline{1.875} & \underline{0.848} \\ 
                      & 48 & \textbf{\color{blue}1.562} & \textbf{\color{blue}0.783} & 1.662 & \underline{0.809} & \underline{1.636} & \underline{0.800} \\
                      & 60 & \textbf{\color{blue}1.445} & \textbf{\color{blue}0.789} & 1.839 & 0.952 & \underline{1.711} & \underline{0.870}\\
\cmidrule(lr){1-8}
\multirow{2}{*}{Electricity} 
                    & 96 & \textbf{\color{blue}0.153} & \textbf{\color{blue}0.239} &  0.162 & 0.246 & \underline{0.154} & \underline{0.241}\\
                    & 192 & \textbf{\color{blue}0.163} & \textbf{\color{blue}0.248} & \underline{0.164} & 0.250 & 0.165 & \underline{0.250} \\ 
                      & 336 & \textbf{\color{blue}0.182} & \textbf{\color{blue}0.268} & 0.186 & \underline{0.271} & \underline{0.184} & 0.272 \\
                      & 720 & \textbf{\color{blue}0.215} & \textbf{\color{blue}0.296} & 0.217 & 0.298 & \underline{0.218} & \underline{0.299}\\
\cmidrule(lr){1-8}
\multirow{2}{*}{Weather} 
                    & 96 & \textbf{\color{blue}0.164} & \textbf{\color{blue}0.202} & \underline{0.181} & \underline{0.207} & \underline{0.181} & \underline{0.207}\\
                    & 192 & \textbf{\color{blue}0.210} & \textbf{\color{blue}0.245} & \underline{0.228} & \underline{0.249} & 
                    0.231 & \underline{0.249}\\ 
                      & 336 & \textbf{\color{blue}0.266} & \textbf{\color{blue}0.286} & 0.287 & 0.293 & \underline{0.282} & \underline{0.289}\\
                      & 720 & \textbf{\color{blue}0.350} & \textbf{\color{blue}0.341} & \underline{0.364} & \underline{0.344} & 
                      0.373 & 0.346\\
\cmidrule(lr){1-8}
\multirow{2}{*}{ETTh1} 
                    & 96 & \textbf{\color{blue}0.376} & \textbf{\color{blue}0.391} & 0.438 & 0.407 & \underline{0.410} & \underline{0.395}
\\ 
                    & 192 & \textbf{\color{blue}0.418} & \textbf{\color{blue}0.419} & 0.537 & 0.458 & \underline{0.489} & \underline{0.438} 
\\ 
                      & 336 & \textbf{\color{blue}0.452} & \textbf{\color{blue}0.439} & 0.539 & 0.460 & \underline{0.515} & \underline{0.452}
\\
                      & 720 & \textbf{\color{blue}0.456} & \textbf{\color{blue}0.466} & 0.633 & 0.522 & \underline{0.545} & \underline{0.488}
\\

\cmidrule(lr){1-8}
\multirow{2}{*}{ETTh2} 
                    & 96 & \textbf{\color{blue}0.284} & \textbf{\color{blue}0.333} &  0.293 & 0.338 & \underline{0.286} & \textbf{\color{blue}0.333}\\
                    & 192 & \textbf{\color{blue}0.355 } & \textbf{\color{blue}0.382} & 0.367 & 0.390 & \underline{0.360} & \underline{0.383} \\ 
                      & 336 & \textbf{\color{blue}0.406} & \textbf{\color{blue}0.418} & 0.420 & 0.427 & \underline{0.413} & \underline{0.420} \\
                      & 720 & \textbf{\color{blue}0.410} & \textbf{\color{blue}0.431} & 0.427 & 0.442 & \underline{0.415} & \underline{0.435}\\
\cmidrule(lr){1-8}
\multirow{2}{*}{ETTm1} 
                    & 96 & \textbf{\color{blue}0.304} & \textbf{\color{blue}0.341} &  0.318 & 0.346 & \underline{0.309} & \underline{0.343}\\
                    & 192 & \textbf{\color{blue}0.352} & \textbf{\color{blue}0.369} & 0.365 & 0.375 & \underline{0.364} & \underline{0.371} \\ 
                      & 336 & \textbf{\color{blue}0.380} & \textbf{\color{blue}0.392} & \underline{0.387} & \underline{0.392} & 0.392 & \underline{0.392} \\
                      & 720 & \textbf{\color{blue}0.436} & \textbf{\color{blue}0.424} & \underline{0.459} & \underline{0.427} & \underline{0.459} & 0.429\\
\cmidrule(lr){1-8}
\multirow{2}{*}{ETTm2} 
                    & 96 & \textbf{\color{blue}0.170} & \textbf{\color{blue}0.250} & \underline{0.175} & \underline{0.258} & 0.181 & 0.259\\
                    & 192 & \textbf{\color{blue}0.237} & \textbf{\color{blue}0.296} & 0.253 & 0.311 & \textbf{\color{blue}0.246} & \underline{0.300} \\ 
                      & 336 & \textbf{\color{blue}0.297} & 0.366 & \underline{0.301} & \textbf{\color{blue}0.337} & 0.306 & \textbf{\color{blue}0.337} \\
                      & 720 & \textbf{\color{blue}0.396} & \textbf{\color{blue}0.392} & \underline{0.402} & 0.397 & 0.407 & \underline{{0.394}}\\
\bottomrule
\end{tabular}}

\end{table}

\subsubsection{Activation function}

Table \ref{tab:comparison_abbelations2} demonstrates the impact of the Fourier-based learnable activation function against the classical softmax-based variant.

\begin{table} [h!]
\centering
\caption{Ablation study of the final activation function. Blue bold results are the best outcomes.}
\footnotesize
\label{tab:comparison_abbelations2}
\begin{tabular}{l|l|c c|c c}
\toprule
\multirow{2}{*}{\textbf{Datasets}} & \multirow{2}{*}{\textbf{Horizon}} 
& \multicolumn{2}{c|}{\textbf{IPatch (Fourrier)}} 
& \multicolumn{2}{c}{\textbf{IPatch (softmax)}}   \\
& & MSE & MAE & MSE & MAE \\
\midrule
\multirow{2}{*}{ILI} 
                    & 24 & \textbf{\color{blue}1.604} & \textbf{\color{blue}\color{blue}0.751} &  1.739 & 0.806 \\
                    & 36 & 1.689 & \textbf{\textcolor{blue}{0.799}} & \textbf{\textcolor{blue}{1.644}} & \textbf{\textcolor{blue}{0.798}}  \\ 
                      & 48 & \textbf{\textcolor{blue}{1.562}} & \textbf{\textcolor{blue}{0.783}} & 1.635 & 0.806 \\
                      & 60 & \textbf{\textcolor{blue}{1.445}} & \textbf{\textcolor{blue}{0.789}} & 1.569 & \textbf{\textcolor{blue}{0.788}} \\
\cmidrule(lr){1-6}
\multirow{2}{*}{Electricity} 
                    & 96 & \textbf{\textcolor{blue}{0.153}} & \textbf{\textcolor{blue}{0.239}} &  0.153 & 0.245  \\
                    & 192 & \textbf{\textcolor{blue}{0.163}} & \textbf{\textcolor{blue}{0.248}} & 0.164 & 0.251  \\ 
                      & 336 & \textbf{\textcolor{blue}{0.182}} & \textbf{\textcolor{blue}{0.268}} & 0.184 & 0.270 \\
                      & 720 & \textbf{\textcolor{blue}{0.215}} & \textbf{\textcolor{blue}{0.296}} & 0.218 & 0.298 \\
\cmidrule(lr){1-6}

\multirow{2}{*}{Weather} 
                    & 96 & \textbf{\textcolor{blue}{0.164}} & \textbf{\textcolor{blue}{0.202}} &\textbf{\textcolor{blue}{0.164}} & \textbf{\textcolor{blue}{0.202}} \\
                    & 192 & 0.210 & \textbf{\textcolor{blue}{0.245}} & \textbf{\textcolor{blue}{0.209}} & \textbf{\textcolor{blue}{0.245}} \\ 
                      & 336 & 0.266 & \textbf{\textcolor{blue}{0.286}} & \textbf{\textcolor{blue}{0.264}} & \textbf{\textcolor{blue}{0.286}} \\
                      & 720 & 0.350 & 0.341 & \textbf{\textcolor{blue}{0.341}} & \textbf{\textcolor{blue}{0.336}} \\
\cmidrule(lr){1-6}
\multirow{2}{*}{ETTh1} 
                    & 96 & \textbf{\textcolor{blue}{0.376}} & \textbf{\textcolor{blue}{0.391}} & 0.379 & 0.391 
\\ 
                    & 192 & \textbf{\textcolor{blue}{0.418}} & \textbf{\textcolor{blue}{0.419}} & 0.438 & 0.428 
\\ 
                      & 336 & \textbf{\textcolor{blue}{0.452}} & \textbf{\textcolor{blue}{0.439}} & 0.477 & 0.443 
\\
                      & 720 & \textbf{\textcolor{blue}{0.456}} & \textbf{\textcolor{blue}{0.466}} & 0.554 & 0.502 
\\
\cmidrule(lr){1-6}
\multirow{2}{*}{ETTh2} 
                    & 96 & \textbf{\textcolor{blue}{0.284}} & \textbf{\textcolor{blue}{0.333}} &  0.285 & \textbf{\textcolor{blue}{0.333}}  \\
                    & 192 & \textbf{\textcolor{blue}{0.355}} & \textbf{\textcolor{blue}{0.382}} & 0.368 & 0.384  \\ 
                      & 336 & \textbf{\textcolor{blue}{0.406}} & \textbf{\textcolor{blue}{0.418}} & 0.412 & 0.421 \\
                      & 720 & \textbf{\textcolor{blue}{0.410}} & \textbf{\textcolor{blue}{0.431}} & 0.413 & 0.432 \\
\cmidrule(lr){1-6}
\multirow{2}{*}{ETTm1} 
                    & 96 & \textbf{\textcolor{blue}{0.304}} & \textbf{\textcolor{blue}{0.341}} &  0.308 & 0.242  \\
                    & 192 & \textbf{\textcolor{blue}{0.352}} & \textbf{\textcolor{blue}{0.369}} & 0.371 & 0.374  \\ 
                      & 336 & \textbf{\textcolor{blue}{0.380}} & \textbf{\textcolor{blue}{0.392}} & 0.392 & 0.393 \\
                      & 720 & \textbf{\textcolor{blue}{0.436}} & \textbf{\textcolor{blue}{0.424}} & 0.450 & 0.428 \\
\cmidrule(lr){1-6}
\multirow{2}{*}{ETTm2} 
                    & 96 & \textbf{\textcolor{blue}{0.172}} & \textbf{\textcolor{blue}{0.254}} &  0.174 & 0.254  \\
                    & 192 & \textbf{\textcolor{blue}{0.237}} & \textbf{\textcolor{blue}{0.296}} & 0.238 & 0.299 \\ 
                      & 336 & 0.297 & 0.366 & \textbf{\textcolor{blue}{0.293}} & \textbf{\textcolor{blue}{0.333}} \\
                      & 720 & 0.396 & 0.392 & \textbf{\textcolor{blue}{0.389}} & \textbf{\textcolor{blue}{0.289}} \\
\bottomrule
\end{tabular}
\end{table}
 
In long-horizon settings, such as ETTh1 at horizon 720, the Fourier-based IPatch achieves an MSE of 0.456 and MAE of 0.466, compared to 0.554 and 0.502 for the softmax variant, representing a substantial improvement of approximately 21\% in MSE. Similarly, on Weather at horizon 720, the Fourier variant yields an MSE of 0.350, slightly higher than the softmax variant’s 0.341, but maintains competitive MAE (0.341 vs. 0.336). These results highlight the Fourier activation’s ability to model nuanced frequency dependencies in the spectral domain, which is critical for long-range forecasting.

For shorter horizons, such as ILI at horizon 24, the Fourier activation achieves an MSE of 1.604 and MAE of 0.751, outperforming the softmax variant (MSE 1.739, MAE 0.806). In cases where performance is comparable (e.g., Weather at horizon 96, with identical MSE of 0.164 and MAE of 0.202), the Fourier activation shows no degradation, demonstrating its robustness across varying forecasting lengths.

 Unlike the softmax activation, which may compress information and lose fine-grained details, the Fourier-based activation excels at capturing frequency-based dependencies, making it particularly suited for datasets with complex temporal patterns (e.g., ETTh1, ETTh2). This is evident in ETTh1 at horizon 192, where the Fourier variant achieves an MSE of 0.418 compared to 0.438 for softmax, indicating better modeling of long-term trends.

\section{Conclusion}
\label{conclusion}

This paper introduces IPatch, a novel Transformer-based framework for time series forecasting. IPatch integrates two key components: patching and a point-wise autocorrelation structure. This hybrid design enables the model to capture local semantic information and leverage long-range temporal dependencies simultaneously, thus addressing one of the central challenges in multivariate forecasting. Unlike previous approaches, which usually prioritize either short-term local dynamics or global structures, IPatch achieves a balanced and complementary modeling of both.

Extensive experiments were conducted on datasets with varying temporal properties, ranging from complex seasonality to linear trends, demonstrating the model's robustness and versatility. IPatch achieves consistent performance gains, with relative improvements of up to 24\% in MSE and 12\% in MAE compared to the strongest baseline, TimeMixer, in diverse datasets and forecasting horizons. The ablation studies provide further insights into the effectiveness of the design. They reveal that the combined interaction between patching and point-wise autocorrelation is central: while the autocorrelation-only variant outperforms the patch-only variant, underscoring the importance of global temporal dependencies, the full integration of both components consistently yields the best results, particularly on complex datasets such as ETTh1 and ETTh2. Moreover, the use of the Fourier-based activation function enhances the model’s ability to capture frequency-dependent structures, leading to substantial improvements over softmax-based formulations, especially for long-horizon forecasting (e.g., up to 21\% reduction in MSE on ETTh1 at horizon 720). 

One promising direction for further research is the development of adaptive patching strategies, where patch lengths and positions are dynamically learned from the data. Such strategies could further enhance the model’s ability to capture both local and global dependencies. Another potential direction is to explore explainability methods that leverage the synergy between patch-based and point-wise autocorrelation components, enabling the extraction of interpretable insights from forecasts and providing actionable guidance to domain experts beyond raw forecasts.


\bibliographystyle{elsarticle-num} 
\bibliography{mainpaper}

\newpage
\appendix
\section{Plots}
\begin{figure}[htbp]
    \centering
    \includegraphics[width=0.48\textwidth]{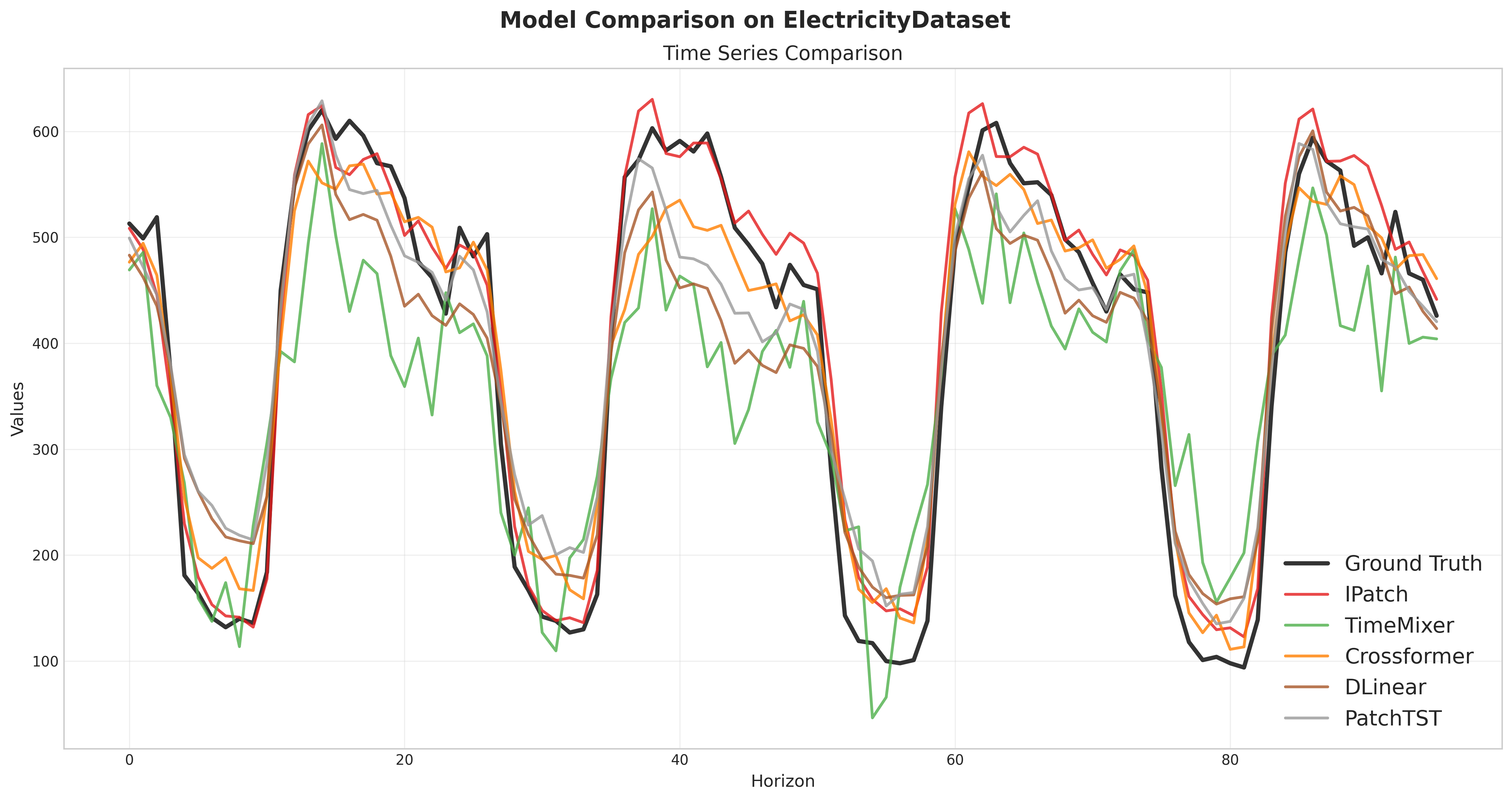}\hfill
    \includegraphics[width=0.48\textwidth]{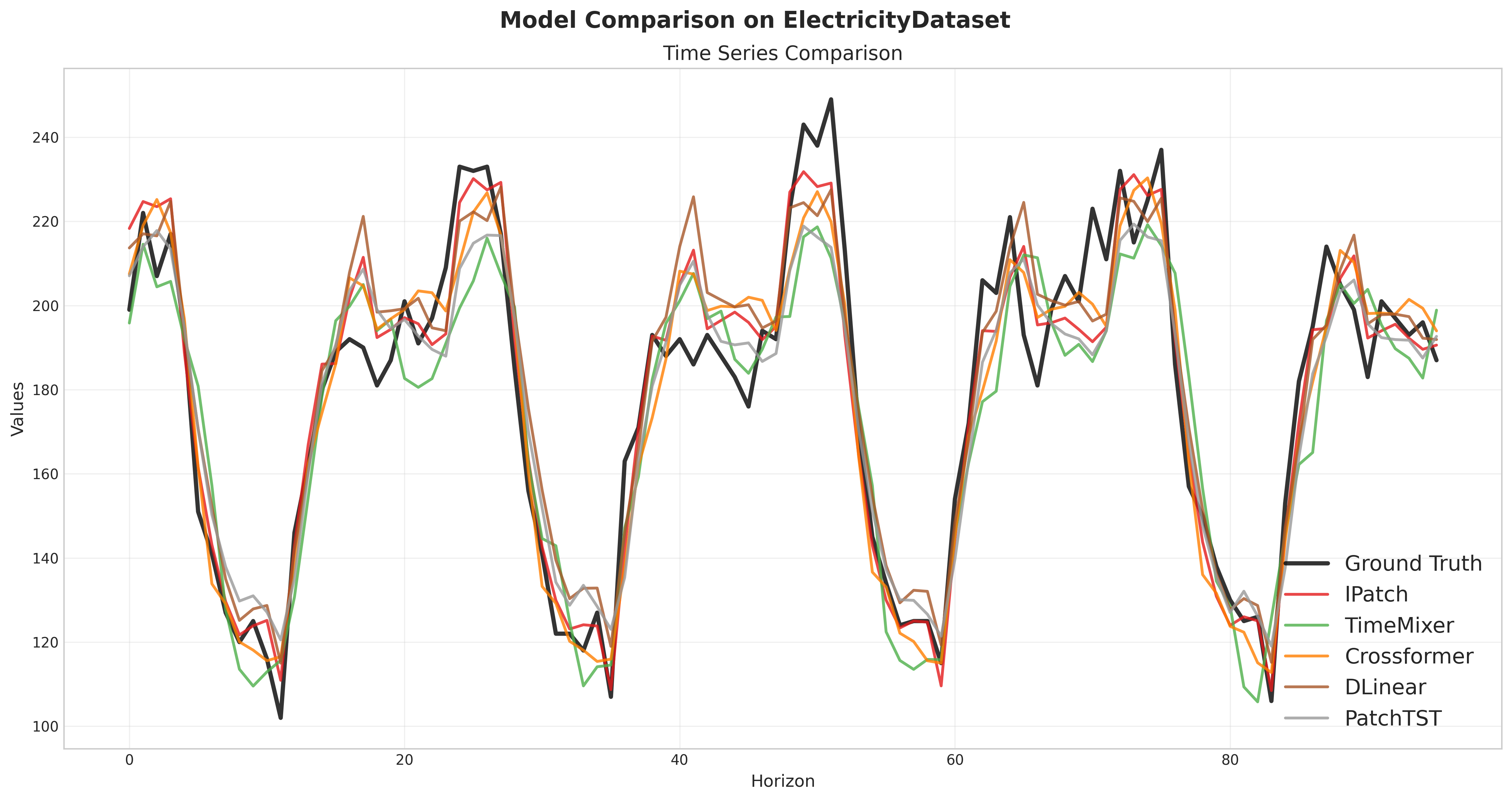}
    \parbox{0.48\textwidth}{\centering (a) Electricity (feature 35)}\hfill
    \vspace{12pt}   
    \parbox{0.48\textwidth}{\centering (b) Electricity (feature 13)}\\[8pt]
    \includegraphics[width=0.48\textwidth]{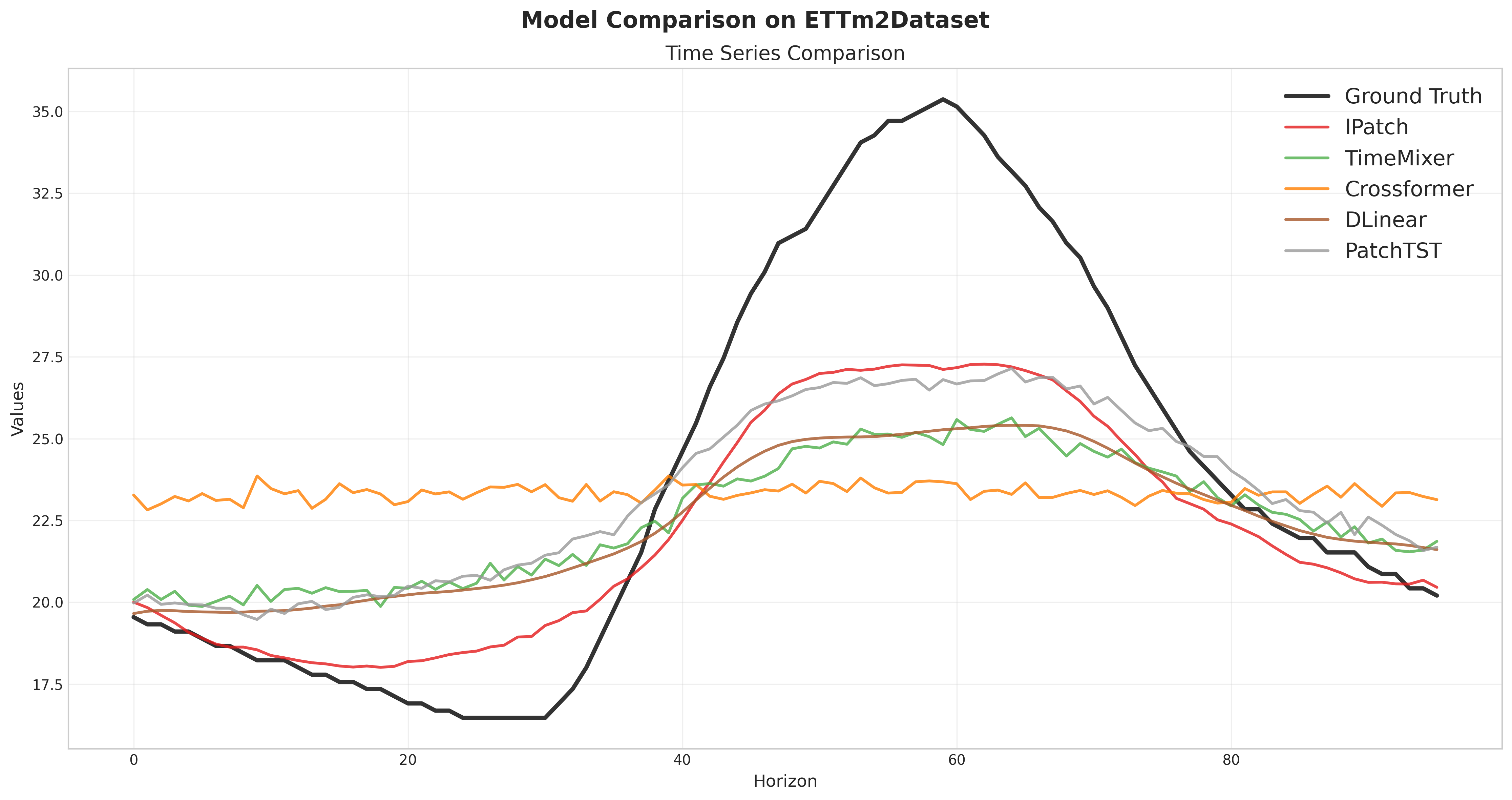}\hfill
    \includegraphics[width=0.48\textwidth]{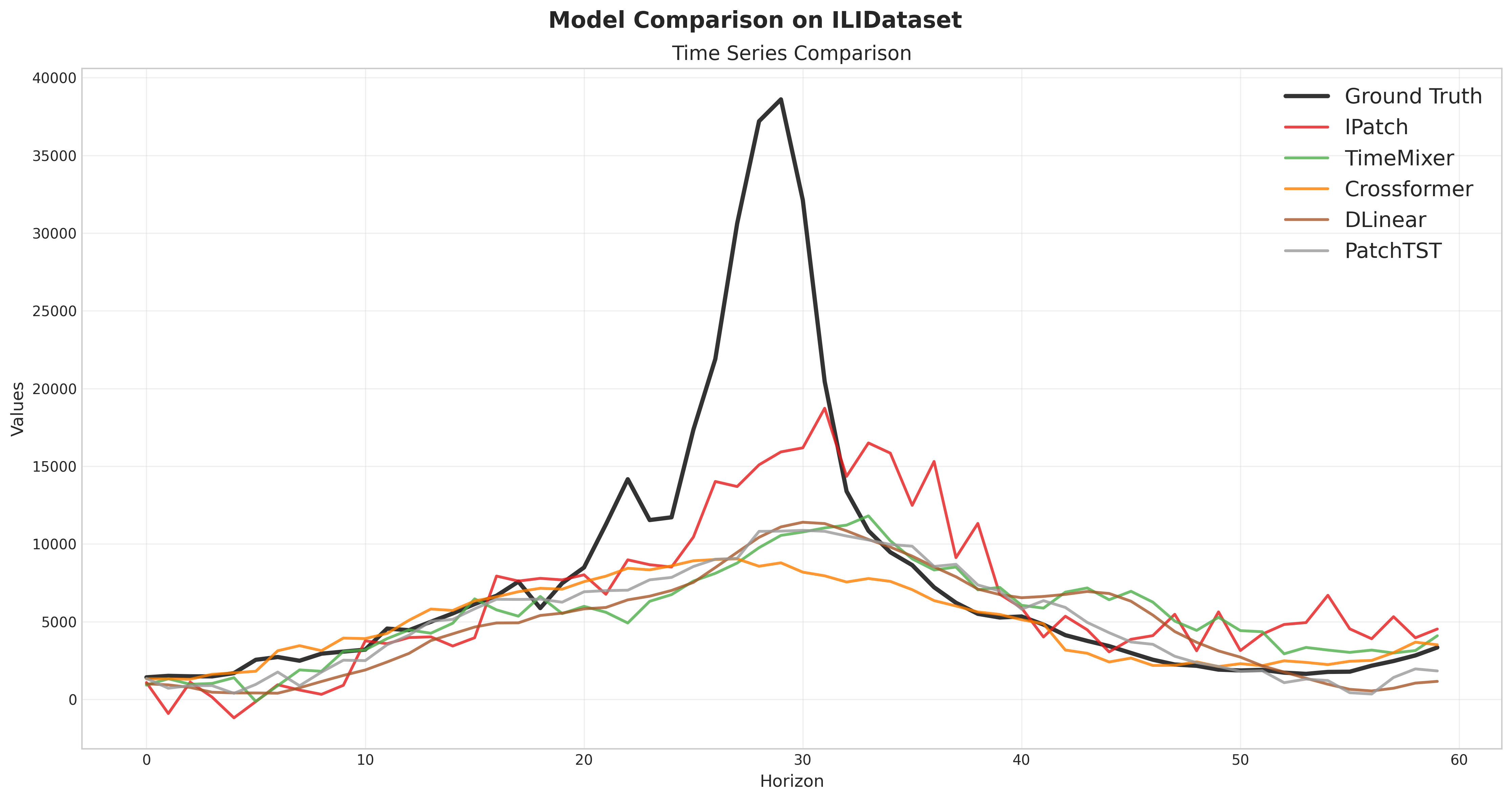}

    \parbox{0.48\textwidth}{\centering (c) ETTm2 dataset}\hfill
    \parbox{0.48\textwidth}{\centering (d) ILI dataset}
    
    \vspace{15pt}
    
    \caption{Qualitative comparison of forecasting models (IPatch, TimeMixer, Crossformer, DLinear, and PatchTST) on different datasets. 
             IPatch shows consistently stronger temporal modeling across diverse time series characteristics.}
    \label{fig:model_comparisons}
\end{figure}

To further demonstrate the robustness of IPatch, Figure~\ref{fig:model_comparisons} presents qualitative forecasting results on representative channels from the Electricity, ETTm2, and ILI datasets, which exhibit markedly different characteristics (strong periodicity, quasi-periodic trends, and spiky non-stationary behavior, respectively). Across all cases, IPatch closely tracks the ground truth with significantly higher fidelity than TimeMixer, Crossformer, DLinear, and PatchTST, especially in capturing sharp peaks, sudden drops, and long-range trends. These visualizations confirm that IPatch delivers consistently superior temporal modeling and generalization capability across diverse time-series regimes.

\end{document}